\documentclass[11pt,a4paper]{article}
\usepackage[hyperref]{acl2021}
\usepackage{times}
\usepackage{latexsym}

\usepackage{microtype}
\usepackage{threeparttablex}

\aclfinalcopy 

\usepackage{subcaption}

\usepackage{multirow}
\usepackage{amsmath}
\usepackage{capt-of}
\usepackage{tabularx}
\usepackage{epsfig}
\usepackage{amssymb}
\usepackage{amsfonts}
\usepackage{booktabs}
\usepackage{scalerel}
\usepackage[inline]{enumitem}
\usepackage{listings}
\usepackage{varwidth}
\usepackage[export]{adjustbox}
\usepackage{tikz}
\usetikzlibrary{tikzmark}
\usepackage{cleveref}

\usepackage{stmaryrd}
\usepackage{bbm}

\usepackage{algorithm}
\usepackage[noend]{algpseudocode}

\definecolor{deepblue}{rgb}{0,0,0.5}
\definecolor{officeblue}{RGB}{0,102,204}
\definecolor{deepred}{rgb}{0.6,0,0}
\definecolor{deepgreen}{rgb}{0,0.5,0}
\definecolor{mybrickred}{RGB}{182,50,28}

\definecolor{fillcolor}{RGB}{216,217,252}

\algnewcommand\algorithmicrequireb{{\hspace{0.85cm}}}
\algnewcommand\INPTDESCB{\item[\algorithmicrequireb]}

\algnewcommand\algorithmicfuncdesc{\textbf{Function:}}
\algnewcommand\FUNCDESC{\item[\algorithmicfuncdesc]}
\algnewcommand\algorithmicfuncdescb{{\hspace{1.48cm}}}
\algnewcommand\FUNCDESCB{\item[\algorithmicfuncdescb]}
\algnewcommand{\algorithmicgoto}{\textbf{goto}}
\algnewcommand{\Goto}[1]{\algorithmicgoto~\ref{#1}}

\usepackage{amsmath,amsfonts,bm}

\def\eqref#1{equation~\ref{#1}}

\def\1{\bm{1}}

\DeclareMathAlphabet{\mathsfit}{\encodingdefault}{\sfdefault}{m}{sl}
\SetMathAlphabet{\mathsfit}{bold}{\encodingdefault}{\sfdefault}{bx}{n}

\newcommand\our{$\Delta$LM}

\title{$\Delta$LM: Encoder-Decoder Pre-training for Language Generation and Translation by Augmenting Pretrained Multilingual Encoders}

\author{
 Shuming Ma, 
 Li Dong,
 Shaohan Huang,
 Dongdong Zhang, \\
 \textbf{Alexandre Muzio,}
 \textbf{Saksham Singhal,}
 \textbf{Hany Hassan Awadalla,}
 \textbf{Xia Song,}
 \textbf{Furu Wei}
 \\
 Microsoft Corporation\\
 \texttt{\{shumma,lidong1,shaohanh,dozhang\}@microsoft.com} \\
 \texttt{\{alferre,saksingh,hanyh,xiaso,fuwei\}@microsoft.com} }

\date{}

\begin{document}
\maketitle
\begin{abstract}

While pretrained encoders have achieved success in various natural language understanding (NLU) tasks, there is a gap between these pretrained encoders and natural language generation (NLG). NLG tasks are often based on the encoder-decoder framework, where the pretrained encoders can only benefit part of it. To reduce this gap, we introduce DeltaLM (\our{}), a pretrained multilingual encoder-decoder model that regards the decoder as the task layer of \textit{off-the-shelf} pretrained encoders.
Specifically, we augment the pretrained multilingual encoder with a decoder and pre-train it in a self-supervised way.
To take advantage of both the large-scale monolingual data and bilingual data, we adopt the span corruption and translation span corruption as the pre-training tasks. Experiments show that \our{} outperforms various strong baselines on both natural language generation and translation tasks, including machine translation, abstractive text summarization, data-to-text, and question generation. The code and pretrained models are available at \url{https://aka.ms/deltalm}.

\end{abstract}

\section{Introduction}

Recently, pretrained language models~\cite{bert,roberta,unilm,t5} have proven effective in many natural language processing tasks. They pre-train a Transformer-based model with self-supervised tasks, and fine-tune it on the downstream tasks, including question answering, sentence retrieval, sentence classification, and so on.

For the natural language understanding (NLU) tasks, the pretrained encoders can initialize most parts except the task layer on the top of the Transformer layers. However, the natural language generation (NLG) tasks are based on the encoder-decoder structure, so the pretrained encoders can only partially benefit them. To eliminate this difference, there are some pretrained language models based on the encoder-decoder architecture, such as BART~\cite{bart,mbart} and T5~\cite{t5,mt5}. They explore some self-supervised tasks to efficiently pre-train the models from scratch.

\begin{figure}[t]
\centering
\includegraphics[width=1.0\linewidth]{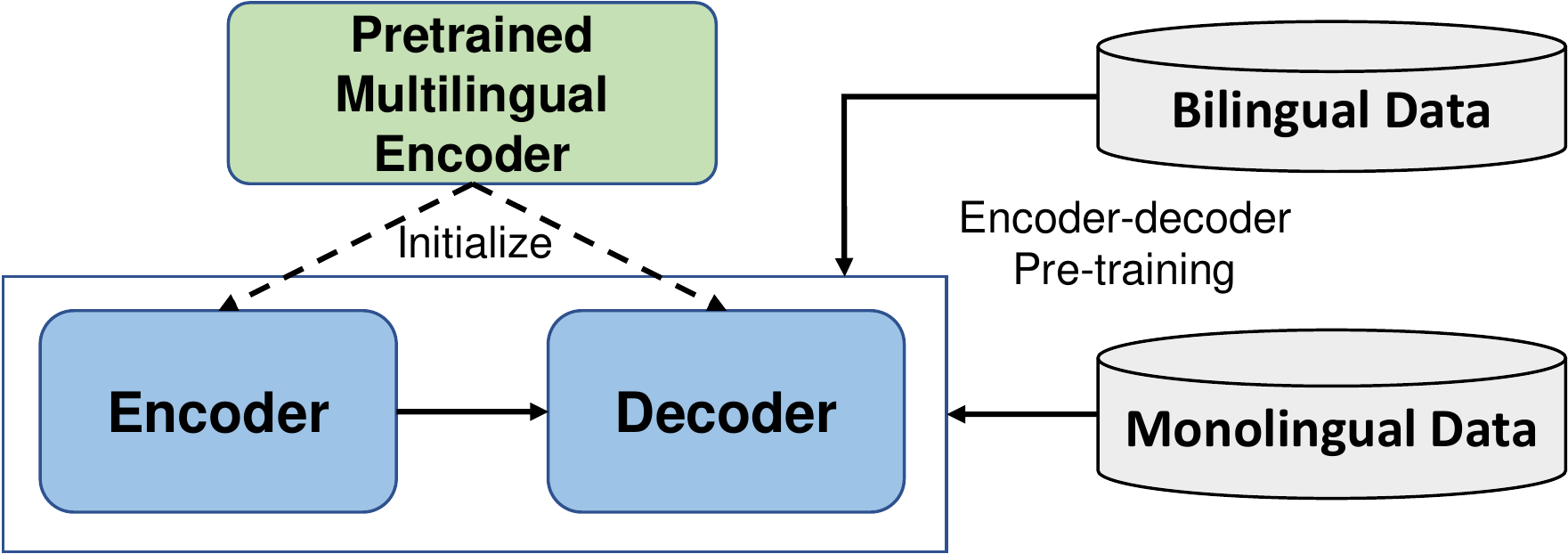}
\caption{Framework of \our{}. We use pretrained multilingual encoders to initialize both the encoder and decoder of the pretrained encoder-decoder model. Then we train it with monolingual data and bilingual data.\label{model}}
\end{figure}

Different from the prior work, we regard the decoder as the 
task layer of \textit{off-the-shelf} pretrained encoders. To achieve this goal, we propose DeltaLM (\our{}), a pretrained multilingual encoder-decoder model, whose encoder and the decoder are initialized with the pretrained multilingual encoder, and trained in a self-supervised way. The overview of \our{} is shown in Figure~\ref{model}.
One challenge is how to initialize the decoder since the architecture of the decoder is different from that of the encoder.
To overcome this problem, we introduce an interleaved decoder that has a more consistent structure with the encoder. 
In this way, the decoder can fully leverage all weights of the pretrained encoder.
The other challenge is which pre-training tasks should be used, as we expect the model can effectively use both large-scale monolingual data and bilingual data.
Inspired by the prior work on pre-training encoder-decoder models, we adopt span corruption~\cite{t5} and translation-pair span corruption~\cite{mt6} as the pre-training tasks. 
The span corruption task proves to be effective in benefiting the cross-lingual transferability from the monolingual data in different languages, while the translation span corruption task helps improve it with the knowledge of bilingual corpora.
Extensive experiments verify the effectiveness of \our{} on both natural language generation and machine translation tasks.


\section{\our{}}

We first initialize both the encoder and the decoder of \our{} with the pretrained encoder. Then, we pre-train \our{} with both monolingual data and bilingual data in a self-supervised way. 

\subsection{Multilingual Pretrained Encoder}

Reusing the pretrained multilingual encoder brings several advantages in terms of both efficiency and effectiveness.
First, it can reduce the training cost by speeding up the convergence. It takes 1 week to train \our{} with 32 V100 GPUs, while mBART~\cite{mbart}, which is trained from scratch, spent 2.5 weeks on training with 256 GPUs.
Second, a strong encoder is important for NLG, according to the empirical studies of the previous work~\cite{deepencoder}.
Our experiments also verify this by comparing the performance of \our{} and the baselines on the downstream tasks.
Third, it can inherit the cross-lingual transferability of the pretrained encoder, which has proven state-of-the-art across various benchmark datasets. 

To take advantage of the strong pretrained multilingual encoder, we use InfoXLM~\cite{infoxlm}. InfoXLM uses the large-scale monolingual data and bilingual data and is jointly trained with a combination of the masked language model, translation language model, and cross-lingual contrast objectives. 
It has a shared vocabulary of 250,000 tokens based on the SentencePiece model~\cite{sentencepiece}.

\begin{figure}[t]
    \centering
    \hspace*{\fill}
    \subcaptionbox{Vanilla decoder\label{fig:classic-decoder}}{\includegraphics[width=0.38\linewidth]{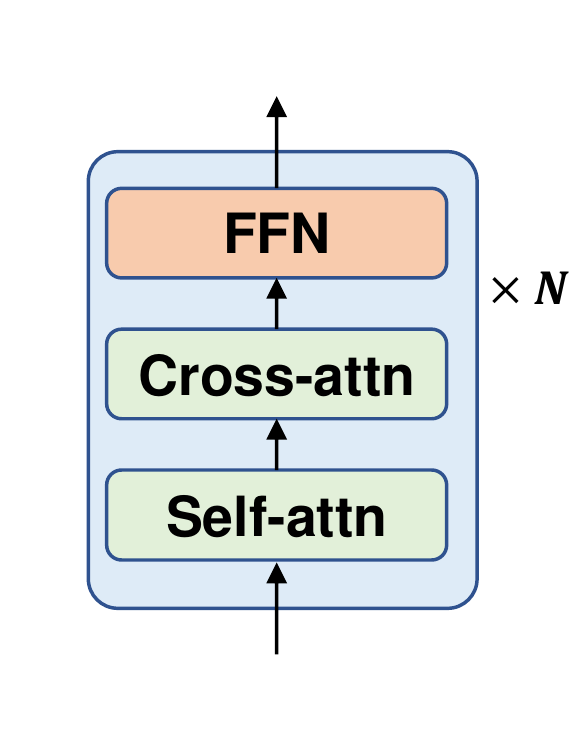}}\hfill
    \subcaptionbox{Interleaved decoder\label{fig:interleaved-decoder}}{\includegraphics[width=0.38\linewidth]{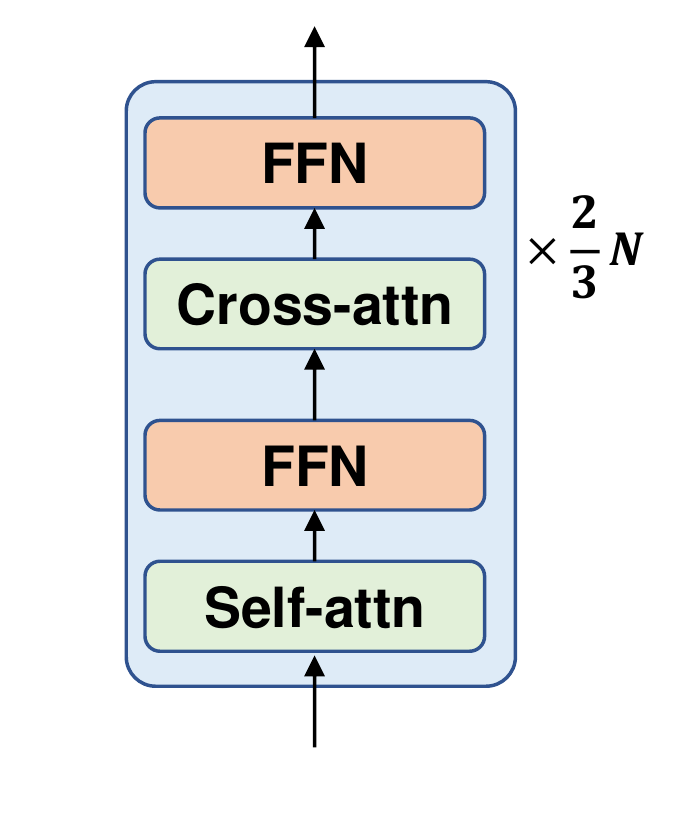}}
    \hspace*{\fill}
    \caption{Illustration of the vanilla Transformer decoder (left) and the proposed interleaved Transformer decoder (right). For simplicity, we omit the embeddings, residual connections, and layer normalization.}
\end{figure}

\subsection{Interleaved Transformer Decoder}

While the encoder can be directly initialized with the pretrained multilingual encoder, it is non-trivial to initialize the decoder, which has a different architecture from the pretrained encoder. Moreover, how to initialize the decoder is under-explored.

As shown in Figure~\ref{fig:classic-decoder}, the vanilla Transformer decoder consists of three modules, including \textit{self-attention}, \textit{cross-attention}, and \textit{feed-forward network (FFN)} in order. Some previous work~\cite{xlm} initializes the \textit{self-attention} and the \textit{FFN} with the weights of pretrained encoder, while the \textit{cross-attention} is initialized with either random weights or the same weights as the \textit{self-attention}.

To better leverage the pretrained encoder, we propose an interleaved Transformer decoder. As shown in Figure~\ref{fig:interleaved-decoder}, we interleave the FFNs and the attention modules, so that the structure is consistent with the pretrained encoder. Then, we replace the each other \textit{self-attention} with a \textit{cross-attention} to maintain the function of the decoder. The residual connections and the layer normalizations are performed in each sub-layers in the same way as vanilla Transformer layers. Therefore, each block consists of one \textit{self-attention}, one \textit{cross-attention}, and two \textit{FFNs}. In this way, the architecture is more consistent with the pretrained encoder.

\paragraph{Initialization} 

With the interleaved structure, we can directly initialize the decoder with the pretrained encoder. More specifically, we initialize the \textit{self-attentions} and the bottom \textit{FFNs} with the odd layers of InfoXLM, while the \textit{cross-attentions} and the top \textit{FFNs} are initialized with the corresponding even layers. We believe the cross-lingual ability of InfoXLM can benefit the \textit{cross-attention}. The rest components of the decoder, including the embeddings, residual connections, the activation function, and the layer normalizations, are the same as the pretrained encoder. Thanks to the interleaved structure, our method can fully use the pretrained weights, and none of the sub-layer should be randomly initialized.

\begin{figure}[t]
    \centering
    \hspace*{\fill}
    \subcaptionbox{Span corruption task\label{sp}}{\includegraphics[width=1.0\linewidth]{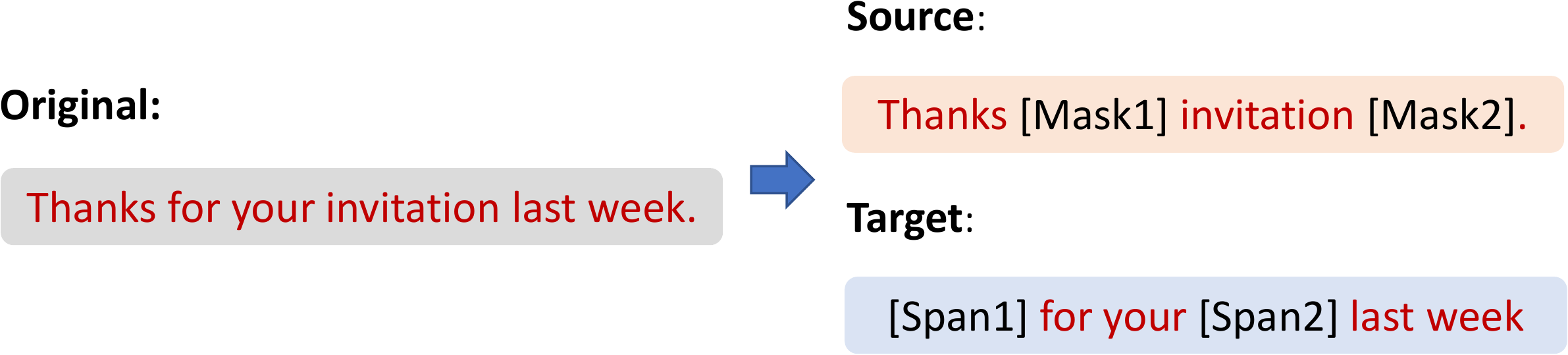}}
    \subcaptionbox{Translation span corruption task\label{tsp}}{\includegraphics[width=1.0\linewidth]{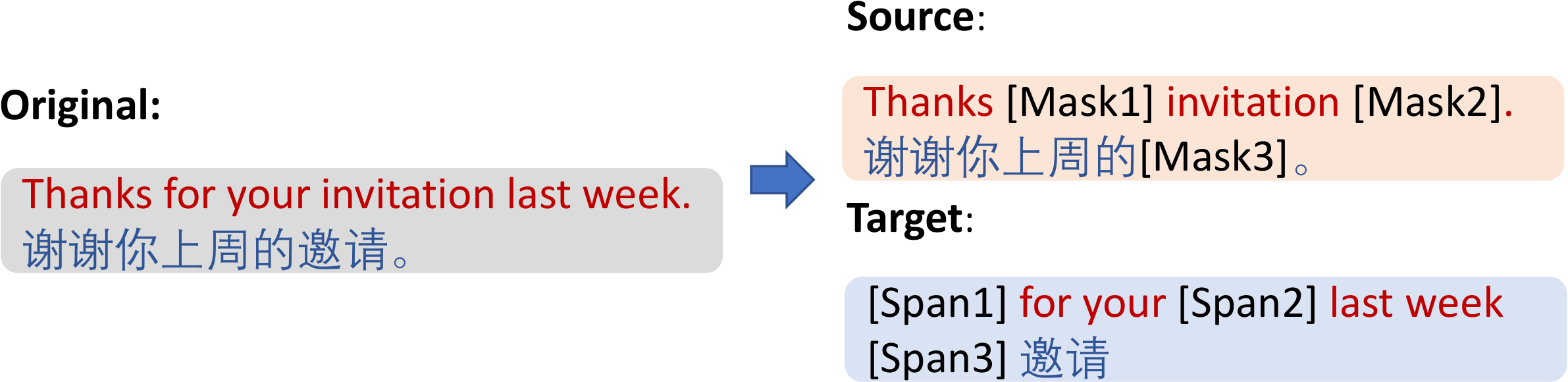}}
    \hspace*{\fill}
    \caption{Illustration of the span corruption task (top) and the translation span corruption task (bottom). Each task constructs an original sample (left) into the text-to-text transform format (right).}
\end{figure}

\subsection{Pre-training Tasks}

After initialization, we pre-train it with two pre-training tasks: span corruption~\cite{t5} and translation span corruption~\cite{mt6}.

\paragraph{Span Corruption}

As shown in Figure~\ref{sp}, span corruption is to reconstruct the text spans based on the masked input document. It is proven to be effective for pre-training an encoder-decoder model. In this work, we follow mT5~\cite{mt5} to apply this pre-training task to pre-train \our{} on large-scale multilingual corpora in 100 languages. We believe this task can help preserve the cross-lingual capability of the pretrained encoders.

\paragraph{Translation Span Corruption}

Since there are available large-scale bilingual corpora, we would like to also leverage this translation data to improve the pretrained encoder-decoder model. Therefore, we follow mT6~\cite{mt6} to improve mT5 with a translation span corruption task. Translation span corruption predicts the text spans based on the input masked translation pair (see Figure~\ref{tsp}). Specifically, we concatenate two parallel sentences as the input and perform the span corruption task. In this way, we can improve the cross-lingual transferability of the model by incorporating the bilingual data.

\paragraph{Pre-training Details}

We pre-train \our{} with 6TB multilingual data, which is a combination of CC100, CC-Net, and Wikipedia, covering 100 languages. We also use 88GB of bilingual data from CCAligned and OPUS, which has 77 languages.
In the experiments, we consider the base-size Transformer model, with $768$ hidden size, $3,072$ FFN dimension, $12$ attention heads, and $12$ encoder/decoder layers. The model is initialized with an InfoXLM \textsc{base} checkpoint~\cite{infoxlm}. We use Adam~\cite{adam} optimizer with $\beta_{1}=0.9$ and $\beta_{2}=0.999$. We adopt a linear learning rate scheduler with $10,000$ warm-up steps. We pre-train the model for $600,000$ steps with $2,048$ samples per batch. For the span corruption task, the input length is 512 tokens, the probability of corrupted tokens is $0.15$ and the average length of spans is $3$. For the translation span corruption, the probability of corrupted tokens is $0.50$ and the span length is $3$. We clip the gradient norm to $1.0$.

\section{Evaluation}

We evaluate \our{} on both natural language generation and translation tasks, including machine translation, abstractive text summarization, data-to-text, and question generation. These various tasks can test the model's capability of multilingual text generation, cross-lingual text generation, and zero-shot cross-lingual transfer.

\begin{table*}[t]
\centering
\small
\begin{tabular}{l|c|ccc|ccc}
\toprule
\multirow{2}{*}{Models} & \multirow{2}{*}{\# Params} & \multicolumn{3}{c|}{XQG-Zh} & \multicolumn{3}{c}{XGiga-Fr} \\
& & BLEU & METEOR & ROUGE-L & BLEU & METEOR & ROUGE-L \\
\midrule
 XLM~\cite{xnlg} & 570M & 23.41 & 23.32 & 47.20 & 56.27 & 39.20 & 52.84 \\
 XNLG~\cite{xnlg} & 480M & 24.89 & 24.53 & 49.72 & 57.84 & 40.81 & 54.24 \\
 \midrule
 \our{} & 360M & \bf 25.80 &\bf  24.87 & \bf 52.05 & \bf 58.39 & \bf 42.02 & \bf 54.94 \\
\bottomrule
\end{tabular}
\caption{Results on Chinese question generation (XQG-Zh) and French abstractive text summarization (XGiga-Fr). \# Params denotes the number of parameters of the model.}
\label{table:monolingual}
\end{table*}

\subsection{Multilingual Language Generation}

We conduct experiments on two tasks of multilingual language generation: abstractive text summarization and question generation, where the source and the target are in the same language.

\paragraph{Abstractive Text Summarization}

Abstractive text summarization is to produce the main points of the input documents with new brief sentences. Following the previous work~\cite{xnlg}, we adopt XGiga as the benchmark dataset. It uses Gigaword~\cite{gigaword} to extract the first sentence and the headline of the articles to construct the document-summary pairs. We use the French XGiga for evaluation. It contains 500k/5k/5k pairs for training/validation/test, respectively. We evaluate the performance by computing the BLEU~\cite{bleu}, METEOR~\cite{meteor} and ROUGE~\cite{lin-2004-rouge} scores.

\paragraph{Question Generation}

Question generation takes an answer and the corresponding passage as the input and generates the related question. 
We use the Chinese XQG as the dataset to evaluate the models. This dataset is constructed from WebQA~\cite{webqa}. We concatenate the passage and the answer into one sequence with a special token \text{[S]} between them. The dataset is split into 135k/5k/3k samples as the training/validation/test sets. The evaluation metrics include BLEU~\cite{bleu}, METEOR~\cite{meteor} and ROUGE~\cite{lin-2004-rouge}.


For both multilingual language generation tasks, we directly fine-tune \our{} on each training set. The optimizer is Adam~\cite{adam} with $\beta_{1}=0.9$ and $\beta_{2}=0.98$. The learning rate is 3e-4 with a warming-up step of 4,000. The models are trained with the label smoothing cross-entropy, and the smoothing ratio is 0.1. The batch size is 4,096 and we accumulate the gradients to simulate a 128-GPU environment.
During testing, we use the beam search algorithm with a beam size of 5 and limit the output sequence to 80 tokens. We select the checkpoints with the best validation performance for all experiments.

\begin{table*}[t]
\centering
\small
\begin{tabular}{l|c|cccccccccc|c}
\toprule
\multicolumn{1}{l|}{X$\rightarrow$En test sets} & \#Params & Fr & Cs & De & Fi & Lv & Et & Ro & Hi & Tr & Gu & Avg \\
\midrule
Bilingual NMT  & 240M\textsuperscript{\textdagger} & 36.2 & 28.5 & 40.2 & 19.2 & 17.5 & 19.7 & 29.8 & 14.1 & 15.1 & 9.3 & 23.0 \\
\midrule
Multilingual NMT & 240M & 34.8 & 29.0 & 40.1 & 21.2 & 20.4 & 26.2 & 34.8 & 22.8 & 23.8 & 19.2 & 27.2  \\
mBART~\cite{mbart} & 610M & 36.2 & 29.9 & 40.0 & 22.2 & 20.6 & 27.2 & 37.2 & 23.3 & 25.7 & 21.7 & 28.4 \\
\midrule
\our{} & 360M & 36.5 & 30.9 & 42.2 & 23.0 & 22.3 & 29.2 & 37.7 & 27.0 & 27.3 & 22.7 & \bf 29.9 \\
\midrule \midrule
\multicolumn{13}{l}{\textit{More training data and language directions}} \\
M2M-100~\cite{m2m100} & 420M & 33.4 & 26.2 & 35.6 & 19.6 & 19.9 & 25.8 & 34.1 & 22.0 & 23.4 & 0.4 & 24.0 \\
M2M-100~\cite{m2m100} & 1.2B & 35.8 & 29.6 & 40.7 & 22.8 & 23.0 & 30.6 & 38.2 & 24.6 & 26.1 & 0.5 & 27.2 \\
\bottomrule
\end{tabular}
\caption{X$\rightarrow$En test BLEU for multilingual machine translation. \textdagger For low-resource languages (Tr, Hi, Gu), we use a small-sized Transformer with 10M parameters to avoid overfitting. \# Params denotes the number of parameters of the model.}
\label{table:x2e}
\end{table*}

\begin{table*}[t]
\centering
\small
\begin{tabular}{l|c|cccccccccc|c}
\toprule
\multicolumn{1}{l|}{En$\rightarrow$X test sets} & \#Params & Fr & Cs & De & Fi & Lv & Et & Ro & Hi & Tr & Gu & Avg \\
\midrule
 Bilingual NMT & 240M\textsuperscript{\textdagger} & 36.3 & 22.3 & 40.2 & 15.2 & 16.5 & 15.0 & 23.0 & 12.2 & 13.3 & 7.9 & 20.2\\
\midrule
 Multilingual NMT & 240M & 34.2 & 20.9 & 40.0 & 15.0 & 18.1 & 20.9 & 26.0 & 14.5 & 17.3 & 13.2 & 22.0 \\
 mBART~\cite{mbart} & 610M & 33.7 & 20.8 & 38.9 & 14.5 & 18.2 & 20.5 & 26.0 & 15.3 & 16.8 & 12.9 & 21.8 \\
\midrule
\our{} & 360M & 35.8 & 22.4 & 40.9 & 15.7 & 18.8 & 20.6 & 26.9 & 17.3 & 18.5 & 16.2 & \bf 23.3 \\
\midrule \midrule
\multicolumn{13}{l}{\textit{More training data and language directions}} \\
 M2M-100~\cite{m2m100} & 420M & 31.5 & 18.4 & 33.9 & 13.1 & 15.4 & 18.6 & 27.9 & 17.3 & 14.5 & 0.3 & 19.1 \\
M2M-100~\cite{m2m100} & 1.2B & 35.5 & 22.1 & 42.2 & 16.6 & 19.2 & 22.9 & 32.0 & 17.9 & 15.5 & 1.3 & 22.5 \\
\bottomrule
\end{tabular}
\caption{En$\rightarrow$X test BLEU for for multilingual machine translation. \textdagger For low-resource languages (Tr, Hi, Gu), we use a small-sized Transformer with 10M parameters to avoid overfitting. \# Params denotes the number of parameters of the model.}
\label{table:e2x}
\end{table*}

\subsection{Cross-lingual Language Generation}

Besides multilingual language generation, we also test the performance on cross-lingual language generation tasks, where the target language is different from the source language. We perform experiments on machine translation, cross-lingual text summarization, and data-to-text generation.

\paragraph{Machine Translation}

As for machine translation, we evaluate the models on the large-scale WMT-10 benchmark dataset~\cite{zcode,xlmt}. This dataset is a collection of parallel data in different languages from the WMT shared tasks. The parallel data is between English (En) and other 10 languages, including French (Fr), Czech (Cs), German (De), Finnish (Fi), Latvian (Lv), Estonian (Et), Romanian (Ro), Hindi (Hi), Turkish (Tr) and Gujarati (Gu). It contains 32.5 million sentence pairs in the training set. We combine all the parallel data in different languages as the training set and evaluate the models on the test sets in each language. We report the case-sensitive detokenized BLEU using sacreBLEU\footnote{BLEU+case.mixed+lang.\{src\}-\{tgt\}+numrefs.1+smooth.exp+tok.13a+version.1.4.14}~\cite{sacrebleu}.

\paragraph{Cross-lingual Text Summarization}

Cross-lingual text summarization aims to generate the summary of the input document in different languages. We adopt WikiLingua~\cite{wikilingua} as the benchmark dataset. It is a large-scale multilingual dataset with about 770k article-summary pairs. The dataset is constructed from WikiHow. Following the previous work~\cite{gem}, we perform experiments in the language directions from Spanish (Es), Russian (Ru), Vietnamese (Vi) and Turkish (Tr) to English (En). The evaluation metric is ROUGE~\cite{lin-2004-rouge}, including ROUGE-1, ROUGE-2, and ROUGE-L. For a fair comparison with the previous work~\cite{gem}, we report the results on the validation set.

\paragraph{Cross-lingual Data-to-text Generation}

Data-to-text generation requires an input of multiple triplets and generates a natural description based on the input data. The benchmark data is WebNLG~\cite{webnlg}. It is a bilingual dataset of parallel DBpedia triple sets and short texts. The language directions are English-English and English-Russian. It contains about 17k triple sets and 45k short texts in English as well as 7k triple sets and 19k texts in Russian. We report the ROUGE scores, including ROUGE-1, ROUGE-2, and ROUGE-L. For a fair comparison with the previous work~\cite{gem}, we report the results on the validation set.


For machine translation, we collect all parallel data in different languages as the training set. To balance the high-resource languages and the low-resource languages, we adopt a dynamic data sampling scheduler~\cite{xlmt}, where the sampling temperature increases from 1.0 to 5.0 gradually. For the cross-lingual text summarization and data-to-text generation, we directly fine-tune \our{}
on each training set in a separate language.

The hyper-parameters during fine-tuning are the same as those used for multilingual language generation. We limit the source length and the target length to 256 for WMT-10 datasets and truncate the inputs to be 512 tokens for WikiLingua. 
During testing, we use the beam search algorithm with a beam size of 5. The last 5 checkpoints are averaged only on WMT-10 for a fair comparison with the previous model.

\begin{table*}[t]
\centering
\tiny
\begin{tabular}{lc|ccc|ccc|ccc|ccc|ccc}
\toprule
\multirow{2}{*}{Models} & \multirow{2}{*}{\#Params} & \multicolumn{3}{c|}{Es} & \multicolumn{3}{c|}{Ru} & \multicolumn{3}{c|}{Vi} & \multicolumn{3}{c|}{Tr} & \multicolumn{3}{c}{Avg} \\
& & R-1 & R-2 & R-L & R-1 & R-2 & R-L & R-1 & R-2 & R-L & R-1 & R-2 & R-L & R-1 & R-2 & R-L \\
\midrule
 mBART~\cite{gem} & 610M & 38.3 & 15.4 & 32.4 & 33.1 & 11.9 & 27.8 & 32.0 & 11.1 & 26.4 & 34.4 & 13.0 & 28.1 & 34.5 & 12.9 & \bf 28.7 \\
 mT5 small~\cite{gem} & 300M & 29.8 & 9.8 & 25.5 & 27.2 & 8.5 & 23.2 & 29.4 & 10.9 & 23.4 & 23.5 & 6.0 & 19.0 & 27.5 & 8.8 & 22.8 \\
 mT5 base~\cite{gem} & 580M & 36.3 & 13.7 & 30.6 & 32.5 & 11.1 & 26.9 & 32.5 & 13.6 & 26.0 & 26.0 & 7.5 & 20.5 & 31.8 & 11.5 & 26.0 \\
 \midrule
 \our{} & 360M & 36.5 & 13.6 & 29.7 & 33.4 & 12.0 & 27.2 & 31.8 & 10.8 & 25.7 & 39.6 & 17.1 & 32.3 & \bf 35.3 & \bf 13.4 & \bf 28.7 \\
\midrule \midrule
 \multicolumn{17}{l}{\textit{Much larger model size}} \\
  mT5 large~\cite{gem} & 1.2B & 39.3 & 15.7 & 33.0 & 35.0 & 12.7 & 28.8 & 29.9 & 9.6 & 23.8 & 36.2 & 15.0 & 29.1 & 35.1 & 13.3 & 28.7 \\
 mT5 XL~\cite{gem} & 3.7B & 41.8 & 17.4 & 34.7 & 38.6 & 15.4 & 32.3 & 35.5 & 13.0 & 29.2 & 41.5 & 19.6 & 34.7 & 37.4 & 16.4 & 32.7 \\
\bottomrule
\end{tabular}
\caption{Results on cross-lingual abstractive summarization (WikiLingua). \# Params denotes the number of parameters of the model. }
\label{table:wikilingua}
\end{table*}

\begin{table*}[t]
\centering
\small
\begin{tabular}{lc|ccc|ccc|ccc}
\toprule
\multirow{2}{*}{Models} & \multirow{2}{*}{\#Params} & \multicolumn{3}{c|}{En} & \multicolumn{3}{c|}{Ru} &  \multicolumn{3}{c}{Avg} \\
& & R-1 & R-2 & R-L & R-1 & R-2 & R-L & R-1 & R-2 & R-L \\
\midrule
 mBART~\cite{gem} & 610M & 83.4 & 63.1 & 70.3 & 34.8 & 13.4 & 33.0 & 59.1 & 38.3 & 51.7 \\
 mT5 small~\cite{gem} & 300M & 78.8 & 59.2 & 67.2 & 29.7 & 10.5 & 28.4 & 54.3 & 34.9 & 47.8 \\
 mT5 base~\cite{gem} & 580M & 82.3 & 62.1 & 69.7 & 33.0 & 12.7 & 31.3 & 57.7 & 37.4 & 50.5 \\
 \midrule
 \our{} & 360M & 83.4 & 63.9 & 71.1 & 35.0 & 15.0 & 33.3 & \bf 59.2 & \bf 39.4 & \bf 52.2 \\
\midrule \midrule
 \multicolumn{11}{l}{\textit{Much larger model size}} \\
 mT5 large~\cite{gem} & 1.2B & 83.8 & 64.4 & 71.6 & 33.4 & 13.4 & 32.1 & 58.6 & 38.9 & 51.9 \\
 mT5 XL~\cite{gem} & 3.7B & 83.5 & 63.6 & 71.0 & 34.3 & 13.7 & 32.8 & 58.9 & 38.7 & 51.9 \\
\bottomrule
\end{tabular}
\caption{Results on data-to-text generation (WebNLG).  \# Params denotes the number of parameters of the model.}
\label{table:webnlg}
\end{table*}

\begin{table*}[t]
\centering
\small
\begin{tabular}{l|c|ccc|ccc}
\toprule
\multirow{2}{*}{Models} & \multirow{2}{*}{\#Params} & \multicolumn{3}{c|}{XGiga-Fr} & \multicolumn{3}{c}{XGiga-Zh} \\
& & ROUGE-1 & ROUGE-2 & ROUGE-L & ROUGE-1 & ROUGE-2 & ROUGE-L \\
\midrule
 XLM~\cite{xnlg} & 570M & 14.53 & 1.80 & 13.43 & 0.71 & 0.28 & 0.70 \\
 XLM+MT~\cite{xnlg} & 570M & 38.48 & 18.86 & 34.98 & 36.96 & 22.03 & 33.99 \\
 XNLG~\cite{xnlg} & 480M & 39.98 & 20.31 & 36.31 & 41.66 & 28.70 & 38.91 \\
\midrule
 \our{} & 360M & \bf 41.42 & \bf 22.24 & \bf 37.99 & \bf 46.37 & \bf 34.34 & \bf 43.85 \\
\bottomrule
\end{tabular}
\caption{Results on zero-shot abstractive summarization. They are trained on an English dataset and evaluated on the French and Chinese test sets. \# Params denotes the number of parameters of the model.}
\label{table:zero-shot}
\end{table*}

\subsection{Zero-shot Cross-lingual Transfer}

Zero-shot transfer is an important ability of the pretrained language model. To test the cross-lingual transferability, we conduct experiments on the zero-shot abstractive text summarization.

\paragraph{Zero-shot abstractive text summarization}

We use the XGiga to perform experiments. In this setting, we train the model on the English-English training set and evaluate it on the French-French and Chinese-Chinese test sets. The training data consists of 50k text-summary pairs, while both the validation and test sets have 5k samples. We evaluate the performance by computing the ROUGE scores.


Direct fine-tuning leads to ``accidental translation''~\cite{mt5} errors in a language unseen. These errors include illustrating normalization, grammatical adjustment, and translation. Inspired by the previous work~\cite{mt5}, we mix the pre-training tasks into the fine-tuning stage. During fine-tuning, half of the time is to fine-tune the downstream tasks while the rest uses the same objective as the pre-training tasks. Specifically, we remove the sentinel tokens from the target sequence in the span corruption tasks to prevent their appearance in the predictions. We use the same languages as the downstream tasks during mixing in the pre-training tasks. The other details of fine-tuning are the same as those used for directly fine-tuning abstractive text summarization as described above.

\section{Results}

We compare \our{} with various state-of-the-art language generation and translation models. We report the results on various generation tasks.

\subsection{Multilingual Language Generation}

We compare \our{} with XLM~\cite{xlm} and XNLG~\cite{xnlg}, two strong baselines for question generation and abstractive text summarization in different languages.

Table~\ref{table:monolingual} summarizes the results on XQG-Zh and XGiga-Fr. \our{} has an improvement of +2.39 BLEU, +1.55 METEOR and +4.85 ROUGE-L over XLM and +0.91 BLEU, +0.34 METEOR and +2.33 ROUGE-L scores over XNLG on XQG dataset.
Besides, it improves XNLG by a significant gain of +0.55/+1.21/+0.70 points on the XGiga dataset.
It concludes that \our{} can achieve consistent improvement over the strong baselines across different metrics, with an even smaller model size.

\subsection{Cross-lingual Language Generation}

For machine translation, we compare \our{} to the state-of-the-art multilingual pretrained language models and multilingual neural machine translation models, including mBART~\cite{mbart} and M2M-100~\cite{m2m100}. We follow the same implementation as~\citet{multilingualfinetune} to fine-tune the mBART model.\footnote{\url{https://github.com/pytorch/fairseq/tree/master/examples/multilingual}} It is noted that mBART has 12 layers with 1024 hidden size, leading to a larger model size than \our{}. We also evaluate the performance of M2M-100 on the same test sets. We use the officially released checkpoints of M2M-100, including the small size model (420M) and the base size model (1.2B).\footnote{\url{https://github.com/pytorch/fairseq/tree/master/examples/M2M_100}}

We evaluate \our{} and these baselines on both many-to-English (X$\rightarrow$En) translation test sets and English-to-many (En$\rightarrow$X) translation test sets.
As shown in Table~\ref{table:x2e} and Table~\ref{table:e2x}, \our{} improves the multilingual NMT model without pre-training by +2.7 average BLEU on X$\rightarrow$En test sets and +1.3 average BLEU on En$\rightarrow$X test sets. Moreover, \our{} outperforms mBART and M2M-100 across 10 languages with fewer parameters.

As for cross-lingual abstractive summarization and data-to-text generation, the baselines include the state-of-the-art pretrained encoder-decoder models, mBART, and mT5. There are different sizes of mT5, including small size (300M), base size (580M), large size (1.2B), and XL size (3.7B). \our{} has 360M parameters, which is smaller than these baselines except mT5 small.

Table~\ref{table:wikilingua} reports the results on the WikiLingua dataset. We compute the average scores of ROUGE-1, ROUGE-2, and ROUGE-L across Es, Ru, Vi, and Tr. It shows that \our{} achieves the scores of 35.3/13.4/28.7 on average. This result is competitive with mT5 large with only 360M parameters while mT5 large has a model size of 1.2B.

Table~\ref{table:webnlg} summarizes the performance on WebNLG dataset. It shows that \our{} outperforms all these baselines, reaching the average scores of 59.2/39.4/52.2 points. It is noted that \our{} achieves better performance than mT5 XL with only 10\% parameters.

\subsection{Zero-shot Cross-lingual Transfer}
We evaluate the zero-shot cross-lingual transferability of \our{} on XGiga dataset, and Table~\ref{table:zero-shot} includes the results of \our{} and the baselines. \our{} is compared to XLM, XLM+MT, and XNLG.
XLM denotes directly fine-tuning the XLM model on the English summarization dataset and test on the Chinese and French test sets. XLM-MT is to translate the French sentences into English with Google Translator before summarizing with XLM and translate the predictions back to French summaries. XNLG is a strong baseline for the zero-shot abstractive summarization, which freezes the decoder of the pretrained model during fine-tuning the English training set.

From the results, we can see that \our{} significantly outperforms all baselines on both French XGiga test sets and Chinese XGiga test sets, which indicates its good capability of zero-shot cross-lingual transfer of language generation.

\section{Related Work}

\paragraph{Encoder-decoder pre-training}

While pretrained encoders~\cite{bert,roberta,unilm} have achieved lots of success for various NLP tasks, pretrained encoder-decoder models are also effective, especially for NLG tasks. T5~\cite{t5} explores different pre-training tasks and proposes to use the span corruption tasks for pre-training. mT5~\cite{mt5} further extends T5 to support the multilingual pre-training. Along this line, mT6~\cite{mt6} improves mT5 by exploring three different text-to-text pre-training tasks and introducing a partially non-autoregressive objective. MASS is also a sequence-to-sequence based pretrained model, which reconstructs a sentence fragment given the remaining part of the sentence~\cite{mass}. There is some work based on the denoising auto-encoder for language model pre-training, including BART~\cite{bart} and mBART~\cite{mbart}. Different from the prior work, our work focuses on reusing the pretrained encoder for encoder-decoder pre-training.

\paragraph{Pretrained multilingual model}

This work is also related to pretrained multilingual model. mBERT~\cite{elements:mbert} extends the BERT model to support different languages with a single pretrained model. XLM~\cite{xlm} explores three pre-training tasks, including mask language model, translation language model, and conditional language model to improve the multilingual pretrained model. XLM-R~\cite{xlmr} takes advantage of both XLM and Roberta~\cite{roberta} to achieve a better performance than XLM. InfoXLM~\cite{infoxlm} proposes a novel cross-lingual contrast to enhance the cross-lingual transferability of the pretrained encoders. There are also other work~\cite{unicoder,veco} to explore different pre-training tasks to further improve the performance of pretrained multilingual model.

\section{Conclusion}

In this work, we introduce \our{}, a powerful pretrained multilingual encoder-decoder model for both language generation and translation. \our{} reuses the state-of-the-art pretrained encoder and can leverage both large-scale monolingual data and bilingual data via encoder-decoder pre-training. Extensive experiments prove the effectiveness of \our{} on various language generation and translation benchmark datasets. In the future, we would like to scale up the model and explore its applications on the NLU tasks.

\bibliographystyle{acl_natbib}
\bibliography{deltalm}

\end{document}